\documentclass{article}
\usepackage{ijcai23}

\usepackage{todonotes}
\usepackage{times}
\usepackage{soul}
\usepackage{url}
\usepackage[hidelinks]{hyperref}
\usepackage[utf8]{inputenc}
\usepackage[small]{caption}
\usepackage{graphicx}
\usepackage{amsmath}
\usepackage{amsthm}
\usepackage{booktabs}
\usepackage{algorithm}
\usepackage[switch]{lineno}
\usepackage{amssymb}
\usepackage{algorithm}
\usepackage{amsmath}
\usepackage{amssymb}
\usepackage{mathtools}
\usepackage{amsthm}
\usepackage{mathabx}
\usepackage[T1]{fontenc}
\usepackage{wrapfig}
\usepackage{subcaption}
\usepackage{caption}
\usepackage{booktabs}
\usepackage{multirow}
\usepackage{algorithm}
\usepackage{amsfonts}
\usepackage{algorithmicx}
\usepackage{algpseudocode}
\usepackage{threeparttable}
\usepackage{tabularx}
\newcolumntype{C}{>{\centering\arraybackslash}X}
\usepackage[normalem]{ulem}

\newcommand{\name}{AutoTruss}
\newcommand{\yw}[1]{\textcolor{red}{(YW: #1)}}

\newcommand{\yc}[1]{\textcolor{blue}{(YC: #1)}}
\newcommand{\jz}[1]{\textcolor{orange}{(JZ: #1)}}
\newcommand{\hide}[1]{}
\newcommand{\FirstStage}{the search stage}
\newcommand{\SecondStage}{the refinement stage}

\title{Automatic Truss Design with Reinforcement Learning}

\author{
Weihua Du$^{1, *}$
\and
Jinglun Zhao$^{1, 4, *}$
\and
Chao Yu$^1$
\and
Xingcheng Yao$^1$
\and
Zimeng Song$^1$
\and
Siyang Wu$^1$
\and 
Ruifeng Luo$^3$
\and
Zhiyuan Liu$^2$
\and
Xianzhong Zhao$^{2, 4}$
\And
Yi Wu$^{1, 4}$
\affiliations
$^1$Institute for Interdisciplinary Information Sciences, Tsinghua University\\
$^2$Tongji University\\ 
$^3$East China Architectural Design \& Research Institute Co. , Ltd. \\
$^4$Shanghai Qi Zhi Institute\\
\emails
$^*$\{duwh20, zhaojl22\}@mails.tsinghua.edu.com\\
}

\begin{document}

\maketitle

\begin{abstract}
Truss layout design, namely finding a lightweight truss layout satisfying all the physical constraints, is a fundamental problem in the building industry.
Generating the optimal layout is a challenging combinatorial optimization problem, which can be extremely expensive to solve by exhaustive search. 
Directly applying end-to-end reinforcement learning (RL) methods to truss layout design is infeasible either, since only a tiny portion of the entire layout space is valid under the physical constraints, leading to particularly sparse rewards for RL training.
In this paper, we develop {\name}, a two-stage framework to efficiently generate both lightweight and valid truss layouts. 
{\name} first adopts Monte Carlo tree search to discover a diverse collection of valid layouts. Then RL is applied to iteratively refine the valid solutions. 
We conduct experiments and ablation studies in popular truss layout design test cases in both 2D and 3D settings. {\name} outperforms the best-reported layouts by 25.1\% in the most challenging 3D test cases, resulting in the first effective deep-RL-based approach in the truss layout design literature. 
\end{abstract}

\section{Introduction}

Truss layout design and optimization is a crucial and fundamental research topic in the building industry, as truss layouts can be found in a wide range of structures, including bridges, towers, roofs, floors~\cite{stolpe2016truss,alhaddad2020outrigger} and even in aerospace and automotive sectors~\cite{wang2019mechanical}. As a basic component in building structures, a truss can support heavy loads and span long distances with a small amount of construction material. Efficient truss layout design can lead to significant cost savings and can also improve the physical performance and safety of the structure.

However, truss layout design is an NP-hard combinatorial optimization problem, which involves the optimization of node locations, topology between nodes, and the cross-sectional areas of connecting bars \cite{fenton2015discrete}. The possible search space for truss layouts is huge, nonlinear, and non-convex. There are also a number of constraints that must be satisfied, including material strength, displacement allowance, and stability of structural members \cite{luo2022alphatruss}. 
Traditionally, engineers design and optimize truss layouts using a combination of mathematical analysis and physical testing based on domain knowledge. They analyze the structural behavior and iteratively adjust the size and shape based on initial sketches~\cite{dorn1964automatic}. These methods rely heavily on subjective human expertise, resulting in a cumbersome and restrictive design process. An automated design approach is crucial for achieving greater efficiency and flexibility in the design process.

Previous studies have attempted to automate the design of truss layouts using heuristic algorithms, such as genetic algorithms \cite{permyakov2006optimum}, particle swarm optimization \cite{luh2011optimal}, simulated annealing \cite{lamberti2008efficient}, and differential evolution\cite{ho2016adaptive}. However, the size and the complexity of the search space impeded the achievement of optimal results. Note that the entire search space, including node positions, is continuous, and a tiny position change may drastically influence the physical performance of the entire truss layout. So, directly applying search-based methods \hide{to this problem} can be particularly expensive. A low-resolution discretization over the search space may easily miss out on the optimal positions and lead to low-quality solutions~\cite{luo2022alphatruss}.

Reinforcement learning (RL) methods have achieved strong results in solving combinatorial optimization problems 
 \cite{mazyavkina2021reinforcement}, such as the Traveling Salesman Problem (TSP)~\cite{bello2016neural} \hide{Mixed Integer Programming (MIP)~\cite{sabharwal2012guiding}}and drug design~\cite{jeon2020autonomous,yoshimori2020strategies}. These problems require the solver to find the optimal combination of a finite set of choices to maximize a certain objective function. Truss layout design is a similar combinatorial problem but it has the following differences. Unlike TSP problems where any order of the cities is feasible, truss layout design and optimization have tight physical constraints, making most truss layouts generated from random actions invalid. This in turn makes reward signals sparse for \hide{the RL agent to learn} RL training. The objective function is also more complex than that in TSP, since there are more performance indices, like capacity and stability, beyond total mass. 
 The settings of truss layout design are more similar to virtual screening in drug design, namely identifying potential drug candidates from large libraries of compounds~\cite{yoshimori2020strategies}. They both have complex constraints and performance indices. However, in drug design, there exists  
 a large amount of data that can be used for pre-training~\cite{jeon2020autonomous}, whereas in truss layout design, little real-world data are available. 
 These facts make it difficult to directly apply end-to-end RL training to truss layout design.

To sum up, the heuristic search methods can generate valid truss layouts, but with sub-optimal quality~\cite{luo2022alphatruss}. On the other hand, RL can produce fine-grained refinement of truss layouts but suffer from sparse rewards. Therefore, we combine them as a two-stage search-and-refine algorithm named {\name}. In {\FirstStage}, we run a search-based method, Upper Confidence bounds applied to Trees (UCT), to derive diverse truss layouts under all physical constraints. In {\SecondStage}, we adopt the SAC algorithm to train an RL policy for refining the valid truss layouts from {\FirstStage}. We conduct experiments in both 2D and 3D cases, and results show that {\name} improved the SOTA performance by $6.8\%$ on average in 2D test cases and \emph{as much as $25.1\%$ in the more challenging 3D test cases}.

\section{Related Work}

\subsection{Truss Representation}

A concise representation of a truss layout is fundamental for \hide{every} the truss layout design, which should capture both geometry and load conditions. There are mainly two types of representations: voxel-based \cite{li2022fluidic}, and graph-based \cite{stolpe2016truss}. We adopt the graph-based method for its accuracy and flexibility. The voxel-based methods divide the design space into small, three-dimensional units called voxels, each assigned a value representing material density~\cite{li2022fluidic,klemmt2023growth,du2018inversecsg\hide{,baron1999voxel}}. These methods cope well with boundary conditions, but cannot accommodate continuous variations in truss topology and is prone to discretization errors.

On the other hand, graph-based methods represent the truss layout as a graph, consisting of coordinates of nodes, bars connecting the nodes, and member area sizes~\cite{fenton2015discrete,stolpe2016truss,lieu2022novel}, but often simplifying it to follow certain grids or only connecting neighboring nodes. Based on a graph-based approach, our method adopts a continuous additive method, allowing for greater flexibility in node connection and truss layout by adding nodes and connections freely from scratch. Furthermore, Graph Neural Network (GNN)~\cite{scarselli2008graph} is well-suited for processing graph-structured data and complex relationships between elements, thus it is widely used in various real-world applications such as social networks~\cite{li2021relevance\hide{,xu2018powerful}}, chemistry~\cite{fung2021benchmarking,yang2021prediction}, and recommendation systems~\cite{guo2020deep,wu2019session}.

\subsection{Truss Design and Optimization}
There have been various methods for truss layout design and optimization over the years. Traditionally, engineers designed truss layouts based on sketches by experience and refined them with analytical math tools \cite{dorn1964automatic}. This empirical method is time-consuming and far from accurate. With the advancement of technology, computer algorithms based on finite element analysis (FEA) have been adopted for faster and more efficient design \cite{mai2021machine}. These algorithms can be divided into two categories: gradient-based and non-gradient-based. Gradient-based algorithms, such as steepest descent, are efficient in converging to a solution but can be complex to implement mathematically and often produces local solutions~\cite{banh2021non,nguyen2018design,banh2019topology,lieu2022novel}.  On the other hand, non-gradient-based algorithms, such as differential evolution (DE) and genetic algorithms (GA), do not require derivative calculations and are more flexible and robust in the presence of multiple local optima. As a relatively new entrant in this category, Monte Carlo Tree Search (MCTS)~\cite{coulom2007efficient} has shown to be highly effective in large search spaces with the success of AlphaGo~\cite{silver2016mastering}, as it balances exploration and exploitation \cite{luo2022alphatruss,luo2022reinforcement}. Different from previous works which simultaneously optimize truss topology and member sizes, we implement a two-stage search-and-refine approach to sequentially optimize topology and member sizes, which greatly reduces the search space and thus improves the training speed as well as the accuracy of the results. In this paper, we adopt UCT~\cite{kocsis2006bandit}, a variant of MCTS, as the search method for deriving various valid truss layouts in {\FirstStage}.

\hide{Similar to \cite{luo2022reinforcement}, we apply MCTS\yc{MCTS or UCT?} to search for initial valid truss layout designs.}

\hide{In most literature, truss generation and optimization are done simultaneously by changing truss topology and member sizes together. However, this can lead to suboptimal solutions as it is difficult to achieve high precision in a vast search space. To overcome this, we implement a two-stage search-and-refine approach, which first searches for suboptimal topology using MCTS and then refines node locations and member sizes with reinforcement learning (RL). This can not only achieve better end result through stage two, but also speeds up the search.}

\subsection{RL for Combinatorial Optimization}

Recently, reinforcement learning (RL) has emerged as a powerful tool for solving challenging combinatorial optimization problems, such as virtual screening in drug design~\cite{wu2019learning,deudon2018learning}. Various RL algorithms have been applied in this field, including value-based methods like Q-learning~\cite{khalil2017learning}, policy-based methods~\cite{bello2016neural} and policy-gradient based methods~\cite{kool2018attention}. One representative method in RL is Soft Actor-Critic (SAC)~\cite{haarnoja2018soft}, which has been used in robotics~\cite{taylor2021active}, autonomous vehicles~\cite{guan2022discrete}, game playing~\cite{zhou2022revisiting} and many others. In this study, we also leverage the power of RL to address a combinatorial optimization problem, which is fine-grain truss refinement. Specifically, we employ SAC algorithm for this task, as it has a high sample efficiency and a strong ability to explore the solution space.

\section{Preliminary}

\subsection{Problem Formulation}
\label{sec: formulation}

The truss layout design task is to minimize the mass of a truss layout by defining node locations, connections between the nodes, and cross-sectional areas of bars. Formally, a truss layout can be represented as a graph $G=(V, E)$, where $V$ is the set of nodes and $E$ is the set of bars. A bar $e \in E$ can be defined as a tuple $e= (u, v, z)$, with nodes $u, v \in V$, and cross-sectional area $ z \in \mathbb{R}$. The mass can be written as
\begin{equation}
    \label{equ:goal}
    \text { Mass }(G) = \sum_{(u, v, z) \in E} z \times\left\|u-v\right\|
\end{equation}
In truss layout design, certain physical constraints need to be satisfied, to ensure displacement, stress, and buckle condition are within capacity, while the length, area, and slenderness of the bars are within the design limit. Constraint details can be found in Appendix A.1. We consider both 2D and 3D settings in this paper. The only difference is the calculation of the bar's cross-sectional area. In 2D settings, the cross-sectional area is only decided by the width of the bar. While in 3D settings, each bar is a hollow round tube, and the cross-sectional area is decided by the outer diameter and its thickness.

\subsection{Upper Confidence Bounds Applied to Trees}

Upper confidence bounds applied to trees (UCT) algorithm~\cite{kocsis2006bandit} modifies Monte Carlo tree search (MCTS) method with Upper Confidence bounds, which searches for the best termination state $s^*$ with the highest reward $R_{UCT}(s^*)$ with a balance between exploration and exploitation~\cite{gelly2007combining}.
Classical UCT is applied to finite states and actions. For each non-termination state $s$, UCT maintains an action-value function $Q(s,a)$  during tree search, which is calculated as Equ.~(\ref{equ:Q_UCT,reward}):
\begin{equation}
    \label{equ:Q_UCT,reward}
    Q(s,a) = \beta W_{\text{mean}}(s,a) + (1 - \beta)W_{\text{best}}(s,a),
\end{equation}
where $W_{\text{mean}}$ denotes the average reward of all the termination states in the subtree rooted at state $s$, and $W_{\text{best}}$ represents the highest reward in the subtree. $\beta$ is a hyper-parameter to control the exploration preference between the average and the best reward~\cite{kocsis2006bandit}.

The policy of UCT $\pi_{UCT}(s)$ selects the action that maximizes the upper confidence bound on the action value by
\begin{align}
    \label{equ:Q_UCT}
    Q_{UCT}(s, a) &= Q(s, a) + c\sqrt{\dfrac{\log n(s)}{n(s, a)}};\\
    \label{equ:pi_UCT}
    \pi_{UCT}(s) &= \text{argmax}_a Q_{UCT}(s, a),
\end{align}
where $n(s)$ is the number of times that state $s$ has been visited, and $n(s, a)$ is the number of times that action $a$ has been taken from state $s$. Whenever a state $s$ is visited, the counter $n(s)$ and $n(s,a)$ will be increased by 1. 

When UCT begins, all the action values will be initialized to 0. In each UCT iteration, the search process starts from the root state $s_0$ and expands the search tree according to Equ.~(\ref{equ:pi_UCT}). Simulation will be executed till a termination state is reached. Then the counters and the action values of visited state-action pairs will be updated accordingly. The process will be repeated within a given budget of search steps.

\subsection{Reinforcement Learning}
Reinforcement learning (RL) trains an agent to learn to make decisions by interacting with an environment and receiving feedback in the form of rewards. The agent's goal is to maximize its total reward over time. 
To apply RL training, we model the problem as a Markov Decision Process (MDP). MDP is parameterized by $\langle S, A, R, P, \gamma \rangle$, where $S$ is the state space, $A$ is the action space, $R$ is the reward function, $P\left(s^{\prime} \mid s, a\right)$ is the transition probability from state $s$ to state $s^{\prime}$ via action $a$, and $\gamma$ is the discount factor. The goal is to find a policy $\pi_\theta$ parameterized by $\theta$ that outputs an action $\pi_\theta(s)\in A$ for each state $s$ and maximizes the accumulative expected reward.
The objective function is shown in Equ.~(\ref{equ:rl_obj}).
\begin{equation}
\label{equ:rl_obj}
    J(\theta)=\mathbb{E}_{a_t\sim\pi_\theta(s_t)}\left[\sum_t \gamma^t R\left(s_t, a_t\right)\right]
\end{equation}


        
    
\subsubsection{Soft Actor-Critic}
Soft Actor-Critic (SAC) is an off-policy reinforcement learning algorithm that combines the actor-critic framework with an entropy term to encourage exploration. 
SAC optimizes  
\begin{equation}
J(\pi) = \mathbb{E}_{\pi}\left[ \sum_t R(s_t,a_t) + \alpha \cdot H\left(\pi(s_t)\right)\right],
\end{equation}
where $ H\left(\pi\right) $ is the entropy of the policy at state $ s_t $, and $\alpha$ is a temperature coefficient balancing exploration and exploitation. 
SAC maintains a data buffer $D$ with all the transition samples and learns a soft Q-function $Q_\psi(s,a)$ parameterized by $\psi$. Assuming the policy is parameterized by $\theta$, SAC optimizes the policy by the following objective
\begin{equation}
J(\theta) = \mathbb{E}_{s_t\sim D}\left[\mathbb{E}_{a_t\sim \pi_\theta(s_t)}\left[ \alpha \log \pi_\theta(a_t|s_t)-Q_\psi(s_t,a_t)\right]\right].
\end{equation}
The temperature $\alpha$ and the parameter $\psi$ of the soft Q-network are also learned similarly.



\begin{figure*}[ht]
    \centering
    \includegraphics[width=0.8\linewidth]{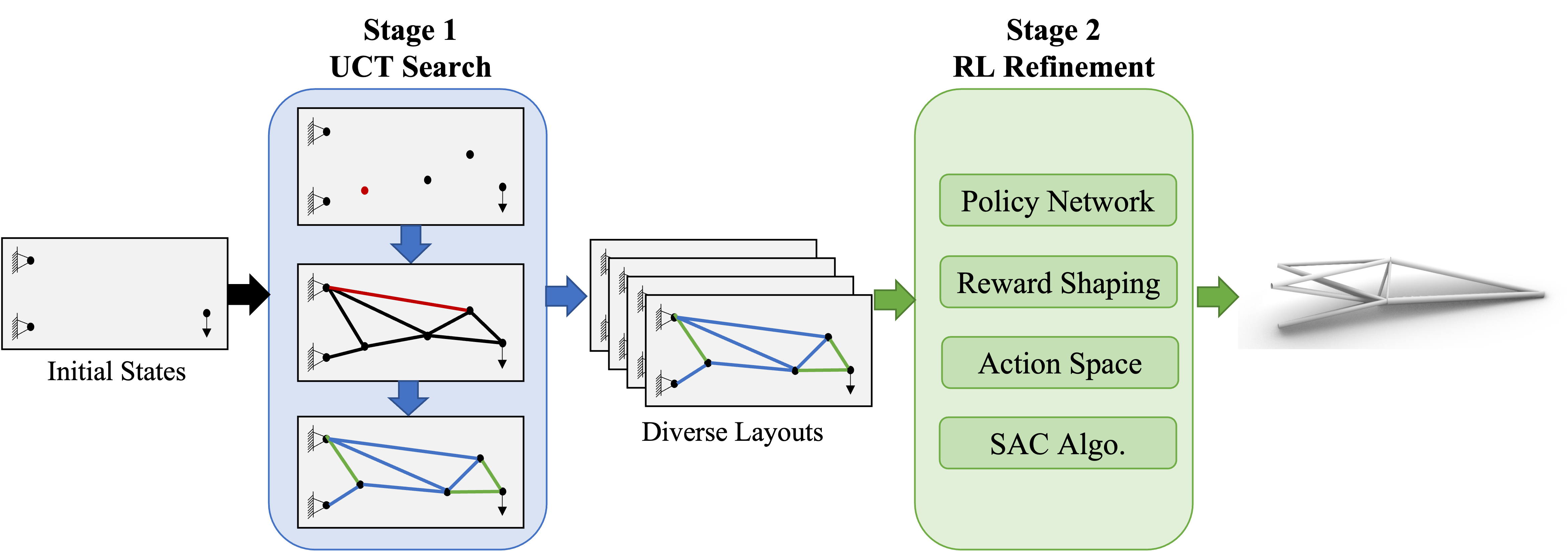}
    \caption{Overview of the two-stage approach {\name}. In {\FirstStage}, UCT Search is applied for diverse valid truss layouts. In the refinement stage, we adopt SAC algorithm to train a policy for truss layout refinement.}
    \label{figure: 3d truss result renderings}  
\end{figure*}

\section{{\name}: A Two-Stage Method}
Truss layout design has a huge search space, which makes it extremely expensive for exhaustive search methods to achieve high performance.  It is not feasible to apply end-to-end reinforcement learning (RL) methods either, since there are many restrictions on valid truss layouts, yielding highly sparse reward signals.  
Therefore, we proposed  {\name}, a two-stage method consisting of a search stage and a refinement stage. In the search stage, {\name} uses a UCT search for diverse \emph{valid} layouts. In {\SecondStage}, {\name} adopts deep RL to further improve the valid solutions. The overview of {\name} is shown in Fig.~\ref{figure: 3d truss result renderings} with details described below. 



\subsection{Search Stage: UCT for Valid Designs}
\label{First Stage}
The purpose of {\FirstStage} is to find diverse \emph{valid} truss layouts as a foundation for {\SecondStage}. We remark that diversity is important since similar topologies from {\FirstStage} will yield similar results from {\SecondStage}, while diverse inputs for the RL policy will improve the overall performances and robustness of {\name}.



\begin{figure}[ht!]
    \centering
    \includegraphics[width=0.9\linewidth]{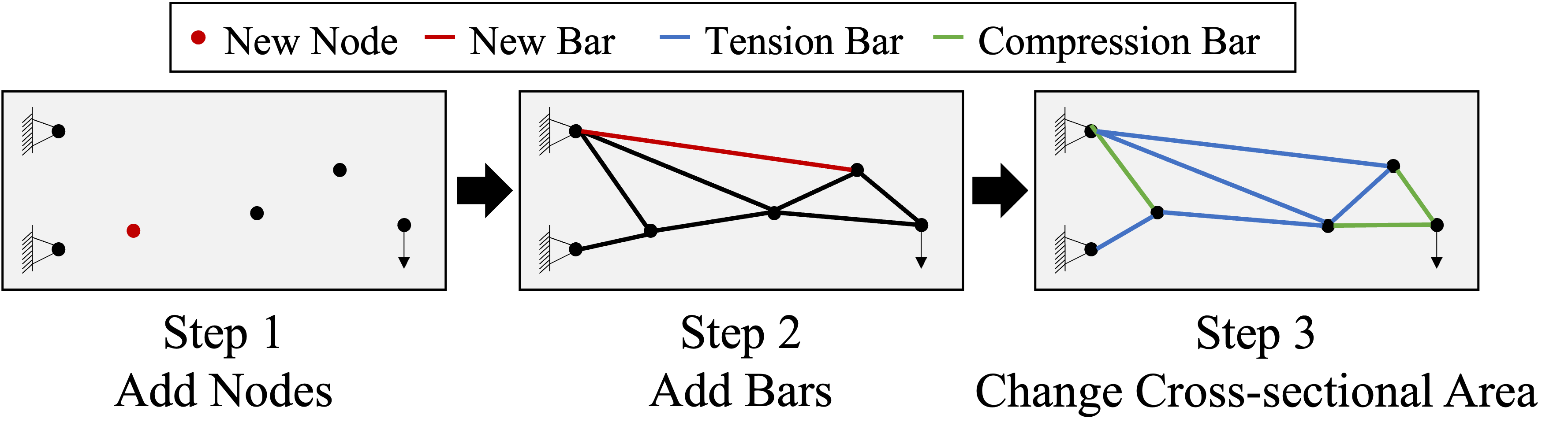}
    \caption{Pipeline of UCT search. UCT search sequentially adds nodes, adds bars, and changes bar cross-sectional area.}
    \label{figure: Stage1}            
\end{figure}


We use UCT search to find valid truss layouts. We divide the generation process of a truss layout into three steps: node-adding step, bar-adding step, and cross-sectional area-changing step. The pipeline of UCT search is shown in Fig.~\ref{figure: Stage1}. To be specific, given the initial truss layout $G_0=(V_0, E_0)$, our UCT search takes these three steps sequentially to produce a complete layout $G_m=(V_m, E_m)$ from scratch. In the node-adding step, it adds new nodes to the layout until it reaches the maximum number of nodes, and then in the bar-adding step, bars with a random cross-sectional area will be added to the truss layout until it satisfies the structural constraints described in Sec.~\ref{sec: formulation}. Finally, we choose the appropriate cross-sectional area for each added bar in the cross-sectional area changing step.
Following \cite{luo2022alphatruss}, for each complete truss layout $G_m=(V_m,E_m)$, the reward is defined by 
\begin{equation}
\label{equ:reward_uct}
    R_{UCT}(G_m) = 
		\left\{
		\begin{array}{cl}
		-1~~, &\text{invalid (structural);}\\
		0~~,  &\text{invalid (other);}\\
		\dfrac{\kappa}{ \text { Mass }(G_m)^2}, &\text{valid layout,}
		\end{array}
		\right..
\end{equation}
$\kappa$ is a scaling parameter, which is typically chosen to bound the maximum reward below 10 for numerical stability.

\hide{UCT search 主要有4个步骤，分别是Selection，Expansion，Simulation，Backpropagation。
Selection：根据 eq(3) 选取在当前节点Q_{UCT}的最大值的儿子节点进行探索。
Expansion：
vanilla UCT仅适用于离散情况。在离散空间中，一个节点的儿子的数量是有限个。因此在扩张新节点时，vanilla UCT可以根据有限个儿子节点数量预先建立搜索树。而在truss design问题中，我们要搜索的节点位置和边的截面积空间是连续的，在连续空间中儿子节点的数量是无限的，所以无法预先建立搜索树，这导致vanilla UCT不可用。为了解决这个问题。
我们引入了KR（cite）来对连续空间中节点的Q uct进行预测。具体来说，在新建一个节点的时候，我们先uniform sample 该节点的25个儿子节点。然后在之后的搜索中再不断加入新的儿子节点。在加入新儿子节点时，我们用已有儿子节点的Q_{UCT}(s, a) 和访问次数 n(s, a)通过kernel regression去预测空间中其他节点的Q_{UCT}(s, a) 和n(s, a)。并根据eq (2)计算所有儿子节点的q uct，将最大q uct值的那个儿子节点加入搜索树。另外，为了控制计算量，我们限制一个节点的访问次数约（小于等于？）等于其儿子节点个数的平方（cite）。
如果走到一个新节点，在UCT search tree 上 expend一层，并进入Simulation环节。
Simulation：到节点后，如果当前桁架结构还没有完成，则random sample 其后继节点，直到桁架结构建立完成，在此过程中不增加新的节点，并得到该桁架结构的reward，该reward只有在桁架结构生成完全结束后才会产生，其定义如下：（公式W（s，a））
Backpropagation：更新所有在路径上的节点的Q和n，具体来说路径上的所有点的W(s, a)_{mean}都会被reward更新，W(s, a)_{max} = max(W(s, a)_{max}, reward), 然后n(s, a) + 1。
}

\subsubsection{UCT Search with Continuous Actions}

A challenge when applying classical UCT to truss layout design is that the actions are all continuous. Therefore, for any intermediate truss layout $G$, finding the optimal UCT action $\pi_{UCT}(G)$ according to Equ.~(\ref{equ:pi_UCT}) becomes non-trivial. In {\name}, we approximate the best action by drawing random samples and choosing the optimal action from the samples:
\begin{align}
    \hat{a}_{(i)}\sim& \text{Uniform}(A)\quad\forall 1\le i\le N,\nonumber\\
    \hat{\pi}_{UCT}(G) =& \arg\max_{a=\hat{a}_{(1)},\ldots,\hat{a}_{(N)}} Q_{UCT}(G, a).
    \label{equ:app_pi_uct}
\end{align}
In our implementation, we choose $N=25$.

Another issue for continuous actions is to compute the action value $Q_{UCT}(G, a)$ since there are infinitely many such values to compute leading to an unbounded search tree size. In our implementation, we constrained the expansion size for each intermediate truss layout $G$ such that we at most expand $O(\sqrt{n(G)})$ children to compute the exact values~\cite{yee2016monte}. For other state-action pair $(G, a')$ without tree expansion, we approximate its $Q(G, a')$ and $n(G, a')$ via kernel regression~\cite{nadaraya1964estimating} based on the precise values of the expanded actions from $G$. Suppose there are $M$ expanded actions, i.e., $\bar{a}_{(1)},\ldots,\bar{a}_{(M)}$. The value $Q(G,a')$ can be approximated by
\begin{align}
\label{equ:kr_uct}
    \hat{Q}(G,a') = \frac{\sum_{i=1}^{M} K(a',\bar{a}_{(i)}) n(G,\bar{a}_{(i)})Q(G,\bar{a}_{(i)})}{\sum_{i=1}^{M} K(a',\bar{a}_{(i)})n(G,\bar{a}_{(i)})}.
\end{align}
The counts $n(G,a')$ can be similarly approximated. 
Here $K(.,.)$ denotes a kernel function. We simply adopt the Gaussian kernel in our implementation. 
\subsubsection{Diverse Layouts}
To get diverse valid truss layouts for {\SecondStage}, we not only need to save the best truss layout, but also some other suboptimal valid truss layouts. Note that two truss layouts $G_1, G_2$ are topologically the same if and only if there exists a permutation $\sigma$ over node indices such that 
\begin{equation}
     \forall (u, v) \in G_1, (\sigma(u), \sigma(v)) \in G_2. 
\end{equation}
It is time-consuming to enumerate all the permutations, we relax the criterion and only adopt the identity permutation in practice for topology checking.
Finally, we store the top 5 lightest valid layouts for each topology and use $\mathcal{G}$ to denote this set of diverse truss layouts we obtained. 



\subsection{Refinement Stage: RL for Adjustment}
\label{Second Stage}
In {\SecondStage}, we adopt the SAC algorithm to refine those \emph{valid} truss layouts $\mathcal{G}$ generated in {\FirstStage}. 


\subsubsection{Action Space}

\hide{\yc{RL policy主要对truss的node位置和edge的截面积进行调整。
对于node的位置的调整是这样的：对于truss的任意一个node，2D case，输出连续动作(\delta_x, \delta_y); 3D case，输出连续动作(\delta_x, \delta_y, \delta_z)，表示该node位置的变化量。 
对于edge的截面积的调整是这样的：对于truss的任意一条边，2D case，输出一个连续动作，i.e. 截面积变化量，表示将该边进行拉伸或者压缩多少。3D case，输出2个连续动作，分别是内外横截面差值和外截面的大小。note that 在3D case，不是所有的截面都是合法的，我们在实际执行的时候会对这两个动作值进行调整，均向上取最小合法值。

另外，每一轮策略更新后，环境本身有一定概率会对架构基于各种constraint再进行合法值检测，并进行调整。}}


The RL policy needs to perform two types of actions, i.e., adjust a node position and the cross-sectional areas of a specific bar in a truss layout.
For node position refinement, when given a specific node to change, the policy outputs a multi-dimensional vector denoting the change of node coordinates. In the 2-dimensional case, the policy outputs $(\delta_x, \delta_y)$ indicating the change in the node's position. Similarly, in the 3-dimensional case, the policy outputs $(\delta_x, \delta_y, \delta_z)$. Here all $\delta_i<0.5$ such that the adjustment will be confined to a small zone with dimension no more than $0.5$m. This is to ensure that the majority of actions taken by RL will not violate the constraints. 
For cross-sectional area changes, when given a specific bar to adjust, the policy outputs a single real value for area change in the 2-dimensional case. 
In the 3-dimensional case, the policy outputs two continuous actions, namely changing the outer diameter and changing the thickness of the bar. Note that not all cross-sectional areas are valid in the 3-dimensional case, so the actual values are rounded up to the minimum legal value during execution.



\hide{In the MDP of stage two for truss optimization, every truss layout state $s$ is described as a graph $G_s = (V_s, E_s)$. There are $3$ types of actions: changing one node $u_i$'s position, changing one bar $e_i$'s cross-sectional area, and checking whether any bar's cross-sectional area can be reduced.\yc{discrete? continuous? or together?}
Specifically, when changing $p_i$'s position, the agent can select a vector $\delta_u = (\delta_x, \delta_y, \delta_z)$, and change $u_i$ into $u_i + \delta_u$. Similarly, when changing $e_i$'s cross-sectional area, the agent can select a vector $\delta_e$ and add it to $e_i$ (The format of $\delta_e$ varies in 2D and 3D cases according to $e_i$).}

\subsubsection{Reward Function}


The design principle of the reward function is to (1) penalize invalid layouts and (2) promote lighter layouts.
Suppose an action $a$ is taken on an intermediate layout $G$ leading to a refined layout $G'$, the reward function is defined as
\begin{equation}
\label{equ:reward1}
    R(G,a) = 
		\left\{
		\begin{array}{rl}
		-50~~~~~~, &\text{invalid (structural);}\\
		-10~~~~~~,  &\text{invalid (other);}\\
		\dfrac{\kappa}{\textrm{Mass}(G')^2} - \dfrac{\kappa}{\textrm{Mass}(G)^2}, &\text{valid.}
		\end{array}
		\right..
\end{equation}
 

\subsubsection{Network Architecture}

\hide{\yc{RL policy的网络架构如图3所示。inspired by transformer（\cite），我们设计了（1）relation encoder对point和edge的位置关系进行提取，（2）action decoder对本轮待操作的point和edge输出高精度的微调动作量。我们用坐标、load、是否为support来表征每个point，用两端的坐标、这条边的截面积(2d)/内外横截面差值和外截面的大小(3d)来表征每条edge。所有的points和edges分别经过embedding层后，送入relation encoder进行关系提取。note that 为了更好的进行关系提取，我们将point-edge连接矩阵作为encoder mask也一起送入relation encoder，point-edge连接矩阵是用1表示点和其相连接的边，其他部分均表示为0。然后，我们将本轮待操作的point和edge信息经过embedding层后的结果，与relation encoder的结果共同送入action decoder。 最后，我们输出两类heads，一个是Q value head，用于critic 进行价值预测；一类是action head，用于输出对待操作point/edge的微调动作量。}}

\begin{figure}[t]
    \centering
    \includegraphics[width=0.9\linewidth]{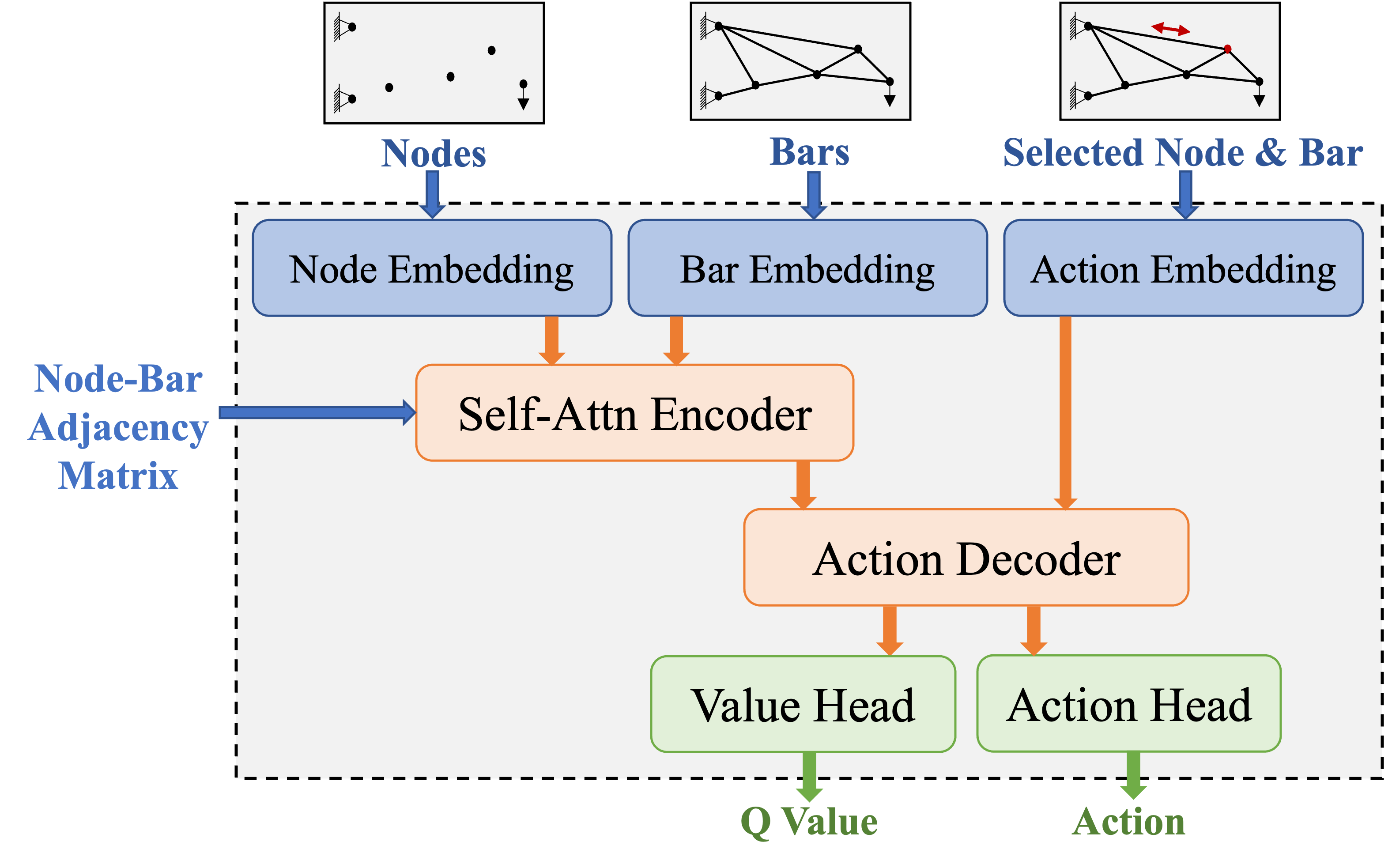}\\
    \caption{Network architecture of RL policy in {\SecondStage}.}\label{figure: RL framework}  
\end{figure}

The network architecture of the RL policy is shown in Fig.~\ref{figure: RL framework}. Inspired by the transformer architecture~\cite{vaswani2017attention}, we adopt (1) a self-attention encoder to extract the spatial relationship between nodes and bars, and (2) an action decoder to output high-precision refinement actions for the node or bar to be operated on in the current iteration. The nodes are represented using coordinates, loads, and whether or not they are supported. The bars are represented as the coordinates of the two end nodes, with (a) the cross-sectional area of the bar in 2D, or with (b) the outer diameter and the thickness of the bar in 3D. All the nodes and bars are passed through an embedding layer and then sent to the self-attention encoder for spatial relationship extraction. The node-bar adjacency matrix is also fed into a self-attention encoder to reflect the topology of the truss layout. 
Then, the results of the embedding layer for the node or bar being operated in the current iteration will be sent to the action decoder together with its embedding of the self-attention encoder. Finally, the policy outputs both the Q values and the multi-dimensional action. 
Full details can be found in Appendix A.4.

\subsubsection{Rollout Generation for RL Training}

In our approach, we employ a probabilistic initialization strategy for the initial state of RL. 
In particular, we keep maintaining the top-5 diverse layouts in $\mathcal{G}$. When each episode starts, we uniformly sample from $\mathcal{G}$ with 50\% probability. Otherwise, we alternatively start from the top-5 lightest truss layouts found during training without considering topology diversity. 
The termination criterion of one episode is that the maximum number of 20 actions are performed. We also early terminate an episode if the policy generates 5 invalid layouts within a single episode. 
In addition, in each RL step, we randomly choose a node or a bar from the current layout for the policy to adjust. 
More details can be found in Appendix A.5.

\hide{In our approach, we employ a probabilistic initialization strategy for the initial state of RL. Specifically, we randomly select one of the five lightest truss layouts generated in {\FirstStage} with a probability of $50\%$, or alternatively, we uniformly select one from the \textit{truss storage} with a probability of $50\%$. The \emph{truss storage} contains the five lightest truss layouts for each truss topology and is updated throughout the optimization process. We terminate the process under two conditions: when the number of actions taken reaches $30$, or when the algorithm generates invalid truss layouts for a total of $5$ times.}


\subsection{Overall Algorithm}

We summarize the overall process of {\name} in Algorithm~\ref{algo:AlphaTruss}. The input to the algorithm is the initial truss layout $G_0 = (V_0, E_0)$ as well as the constraints. $V_0$ represents the support nodes and $E_0$ represents the fixed bars. After applying the two stages, the algorithm finally outputs the lightest truss layout ever derived during the entire search process.

\begin{algorithm}[t!]
\caption{\name}  
\label{algo:AlphaTruss}
\begin{algorithmic}[1]
\algrenewcommand\algorithmicrequire{\textbf{Inputs:}}
  \Require{Initial truss layout $G_0 = (V_0, E_0)$, where $V_0$ means fixed node set and $E_0$ means the fixed bar set.}
\State{Diverse Truss Set $\mathcal{G} \gets\emptyset$}
\While{Tree Search Steps $<$ Limit}
\Comment{search stage}
    \State{Search from $(V_0,E_0)$ w.r.t. Equ.~(\ref{equ:app_pi_uct})}
    \State{Update counts and value for expanded nodes}
    \State{Update $\mathcal{G}$}
\EndWhile
  \State{Initialize the policy $\pi$, data buffer $D$}
\Comment{refinement stage}
  \While {RL steps $<$ RL Limit}
    \State{Select initial state from $\mathcal{G}$}
    \State{Generate an episode $\tau$ w.r.t. the policy $\pi_\theta$}
    \State{Update $\mathcal{G}$}
    \State{Add $\tau$ to D and update $\pi_\theta$ via SAC}
  \EndWhile
\State{\Return{$\arg\min_{G\in\mathcal{G}}\textrm{Mass}(G)$}}
\end{algorithmic}  
\end{algorithm}

\section{Experiments}

We compare {\name} with 3 search-based baselines using both 2D and 3D test cases, where {\name} consistently produces the best truss designs. 
We also evaluate the effectiveness of each module in {\name} through ablation studies. 
We introduce test cases in Sec. \ref{sec: Testbeds}, baselines in Sec. \ref{sec: Baselines}, and the experiment setup in Sec. \ref{sec: Experiment Setup}.
Main results and ablation studies are in Sec. \ref{sec: Main Results} and Sec. \ref{sec: Ablation Study} respectively. \hide{then we conduct ablation studies on 2d cases. Results are over $3$ seeds using a single desktop machine with 1 NVIDIA GeForce RTX 3070 GPU. Full details can be found in Appendix \ref{app...}.} \hide{$\checkmark$\yw{Be precise. We present the main results in sec xxx; then we conduct ablation studies on 2d cases. Results are over xx seeds using a single desktop machine with 1 A100 GPU. Full details can be found in appendix XXXX.}

\yw{How many GPU? how many seeds? how many baselines? how many test cases? from where? More details in appendix. }}

\subsection{Testbeds}
\label{sec: Testbeds}
 \hide{$\checkmark$\yw{meaningless. delete} The objective of truss layout design is to generate structurally sound truss layouts that are as lightweight as possible within the design domain, given a set of fixed nodes, load conditions, and other design constraints.}

\subsubsection{2D Testbed}
\begin{figure}[t]
	\centering
	\includegraphics[width=0.9\linewidth]{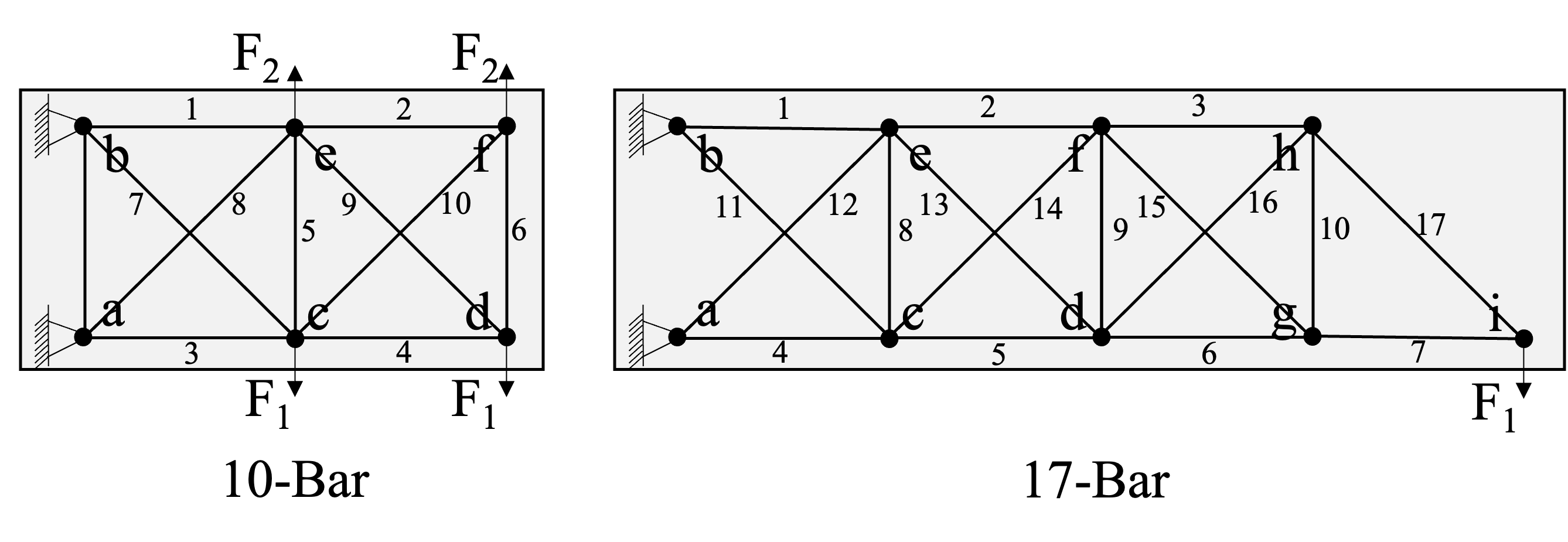}
	\caption{10-Bar and 17-Bar truss layout design cases of the 2D testbed. In the 10-Bar case, there are $2$ kinds of load cases. Load case \uppercase\expandafter{\romannumeral1} has four fixed nodes $(a, b, c, d)$, whereas load case \uppercase\expandafter{\romannumeral2} has six fixed nodes $(a, b, c, d, e, f)$. In the 17-Bar case, there are $3$ fixed nodes $(a, b, i)$. In all 2D cases, $(a, b)$ are support nodes.}
 \label{fig: 2Dtestbed}
\end{figure}


We choose two common 2D test cases in truss layout design~\cite{fenton2015discrete}: the 10-Bar Cantilever Truss (10-Bar) and the 17-Bar Cantilever Truss (17-Bar), as shown in Fig.\ref{fig: 2Dtestbed}. Both are common test cases in the field of structure generation and optimization~\cite{assimi2017sizing,deb2001design,tejani2018size,fenton2015discrete}
\hide{\yw{citation?? how can you make the claim that they are popular?}}. There are $2$ load cases \hide{\yw{change it to 3 different configurations. even two are fine. }} in the 10-Bar case. 
The buckle constraint and the slenderness constraint are not applied to the 10-Bar case. In the 17-Bar case, 
all the constraints except the slenderness constraint are taken into consideration. 
The detailed settings are listed in Appendix A.1. 

\hide{Many heuristic search algorithms based on the ground structure approach test the performance of the algorithm through the 10-bar and 17-bar examples \cite{luo2022alphatruss}\cite{luo2022reinforcement}\cite{fenton2015discrete}\cite{luh2011optimal}. The detailed setting is listed in Appendix \ref{app: 2D setting parameter}.}

\begin{figure}[t]
    \centering
    \includegraphics[width=0.9\linewidth]{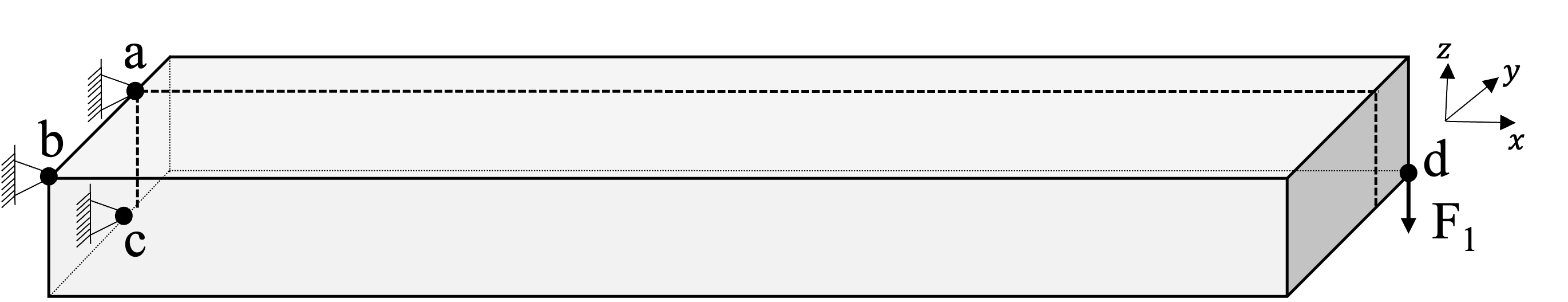}
    \caption{Cantilever Sundial truss layout design case of the 3D testbed. There are $4$ fixed nodes $(a, b, c, d)$. $(a, b, c)$ are support nodes.}
  \label{figure: 3Dtestbed}
  \vspace{-0mm}
\end{figure}

\subsubsection{3D Testbed}
We select the Cantilever Sundial Design (Sundial) as the 3D testbed, which follows \cite{luo2022reinforcement}. The test case was adapted from the sundial bracket truss located in Paternoster Square, London, UK \cite{Shea2004Sundial}. As shown in Fig.~\ref{figure: 3Dtestbed}, the Sundial testbed is characterized by a higher dimension and a larger scale and complexity of the solution space when compared to the 2D testbed. 


\subsection{Baselines}
\label{sec: Baselines}

We consider 3 competitors: \textit{AlphaTruss} \cite{luo2022alphatruss}, \textit{KR-UCT} \cite{luo2022reinforcement}, and \textit{SEOIGE} \cite{fenton2015discrete}.
All the baseline methods can be applied to the 2D testbed, but only \textit{KR-UCT} can be applied to the 3D testbed. Therefore, we compare 2D results with all three baselines and compare 3D results only with \textit{KR-UCT}. We utilized the results of the baselines as reported in their original papers, as the test cases and evaluation methods used in those studies were consistent with those employed in our own research. The details of baselines are listed in Appendix A.2.

\subsection{Experiment Setup}
\label{sec: Experiment Setup}
\textit{KR-UCT} and \textit{SEOIGE} use single-stage search while we use a two-stage search scheme. We balance the iterations in {\FirstStage}, and the environment steps in {\SecondStage} to keep a fair comparison. More specifically, we run $2e6$ iterations in the search stage, which is half the number of iterations in \textit{KR-UCT}, and $1.5e5$ environment steps for RL training, so that the running time of the refinement stage is similar to the search stage with an RTX 3070 GPU. We remark that {\name} consumes substantially fewer trials (i.e., search iterations + RL steps) compared with baselines, and The details can be found in Appendix A.8. We run $3$ seeds for each test case and report the best numbers with the mean numbers and standard deviations. 




\subsection{Main Results}
\label{sec: Main Results}
\subsubsection{2D Results}
\label{sec: 2D results}

\begin{figure}[t]
		\centering
		\includegraphics[width=0.80\linewidth]{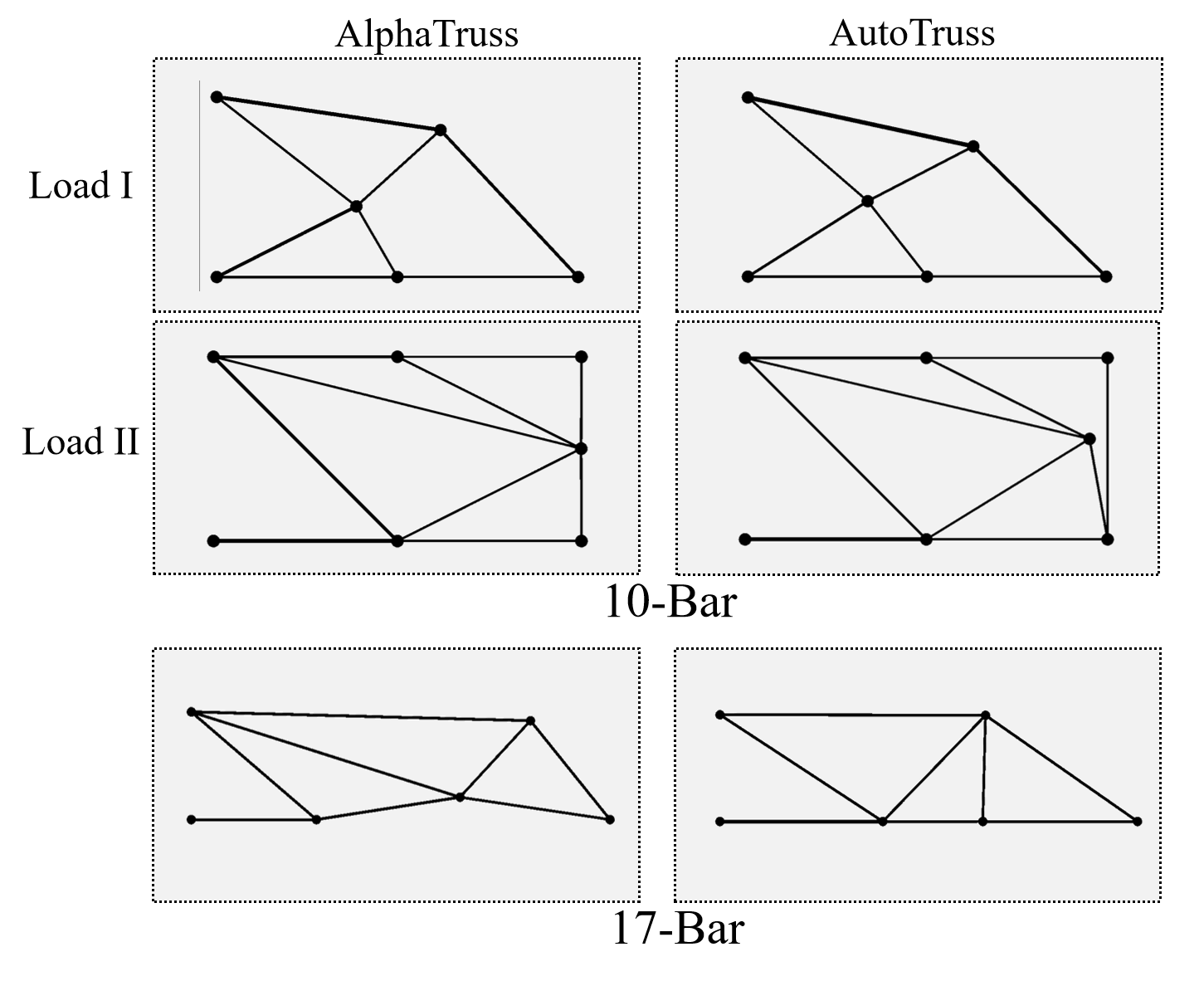}
		\caption{Visualization of truss layouts derived by {\name} and \textit{AlphaTruss} in 2D testbed. We demonstrate 10-Bar truss layouts under 2 settings and the 17-Bar truss layout. The thicker lines represent larger cross-sectional areas. {\name} derives the same truss layout topology as \textit{AlphaTruss} but with better refinement in 10-Bar load \uppercase\expandafter{\romannumeral1} case, and derives better topologies in other cases. The pictures of \textit{AlphaTruss} are adopted from the original paper.}
        \label{fig: 2D_case_visual} 
        \vspace{-2mm}
	\end{figure}

The mass of the solutions derived by {\name} and baselines are presented in Tab.~\ref{tab: 2D_test_case}, where $p$ denotes the number of nodes in truss layouts.
{\name} outperforms the baselines in all the settings by an average of $6.8\%$, demonstrating its ability to discover more lightweight truss layouts. Moreover, The illustration of the truss layouts is shown in Fig.~\ref{fig: 2D_case_visual}. {\name} find the same truss layout topology with better refinement in 10-Bar load \uppercase\expandafter{\romannumeral1} case, and finds better topologies in other cases.  




\begin{table}[t]
		\centering
		\scriptsize
        \setlength{\tabcolsep}{4.5pt}
		\begin{tabular}{cccccc}
		\toprule
		 Cases & Settings & AlphaTruss & KR-UCT & SEOIGE & {\name} \\
		 \midrule
		\multirow{2}{*}{10-Bar} & Load \uppercase\expandafter{\romannumeral1},  $p$=6 & 2150 & 2154 & 2218 & \textbf{2114(2128, 17.6)} \\
		& Load \uppercase\expandafter{\romannumeral2},  $p$=7 & 1616 & N/A & 2098 & \textbf{1337(1410, 61.2)} \\
		 \midrule 
        17-Bar & $p$=6 & 1408 & 1463 & 2582 & \textbf{1378(1398, 22.2)} \\
		 \bottomrule
		\end{tabular}
		\caption{Results of 10-Bar and 17-Bar truss layout design in 2D testbed. $p$ is the number of nodes in the generated truss layouts. N/A denotes that the original paper does not report the number. {\name} outperforms baselines in all cases, showing the capacity to generate lighter truss layouts under various settings.}
        \label{tab: 2D_test_case}
	\end{table}
 
\subsubsection{3D Results}

Tab.~\ref{tab: 3D_experiment_1} shows a comparison of {\name} and \textit{KR-UCT} in the Cantilever Sundial truss layout design of the 3D testbed. Our method consistently outperforms \textit{KR-UCT} by at least $25\%$ under all settings. This highlights the effectiveness of {\name} in designing lightweight truss layouts within a larger search space. It is noteworthy that 3D truss design poses a greater challenge than 2D truss design, as the search space is substantially enlarged. Our approach exhibits a more significant improvement in the 3D case than the 2D counterpart. 

\begin{figure}[t]
		\centering
		\includegraphics[width=0.9\linewidth]{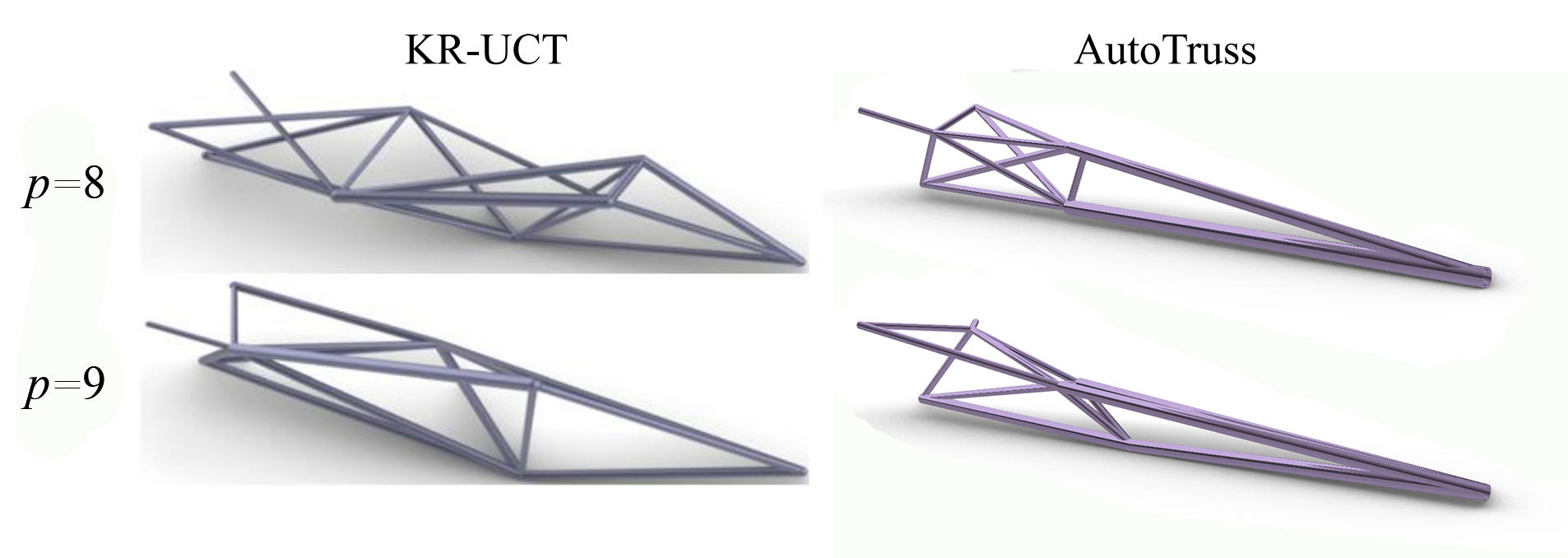}
		\caption{Visualization of truss layouts derived by {\name} and \textit{KR-UCT} in 3D testbed. $p$ is the number of nodes in the generated truss layouts. The truss layouts derived by {\name} are more slender and streamlined than those derived by \textit{KR-UCT}.}\label{figure: 3D Experiment result}.
\end{figure}

The visualization comparison is shown in Fig.~\ref{figure: 3D Experiment result}. The truss layouts derived by {\name} show a more elongated appearance compared with those derived by \textit{KR-UCT}.


\begin{table}[t]
	\centering
	\scriptsize
	\begin{tabular}{cccccccccccc}
	\toprule
	 Settings & KR-UCT & {\name} \\
	 \midrule
	 $p$ = 7 & N/A & \textbf{30.6(31.3, 0.63)} \\
	 $p$ = 8 & 38.7 & \textbf{29.0(30.4, 1.01)} \\
	 $p$ = 9 & 37.2 & \textbf{28.8(30.5, 1.32)} \\
	 \bottomrule
	\end{tabular}
	\caption{Results of Cantilever Sundial truss layout design in 3D testbed. $p$ is the number of nodes in the generated truss layouts. N/A denotes the original paper does not report the number. {\name} outperforms \textit{KR-UCT} by $25.1\%$, showing the ability to generate complex 3D truss layouts.}
	\label{tab: 3D_experiment_1}
\end{table}

\subsection{Ablation Study}
\label{sec: Ablation Study}
In this section, we analyze the effectiveness of the two-stage scheme, the usage of diverse truss layouts in {\FirstStage}, as well as network architecture, all based on the 10-Bar truss layout design cases of 2D testbed through ablation studies. Results are reported as ``mean (standard deviation)''.

\subsubsection{Search-Stage-Only v.s. Two-Stage}

\begin{figure}[t]
    \centering
    \includegraphics[width=0.80\linewidth]{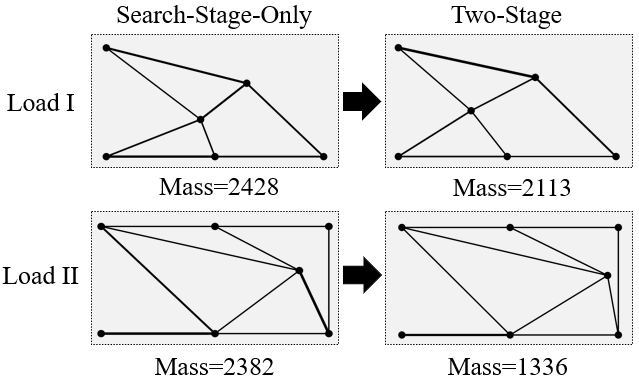}
    \caption{Comparison of truss layouts only derived by {\FirstStage} (\textit{Search-Stage-Only}) and refined by {\SecondStage}(\textit{Two-Stage}). The refinement stage can effectively tune truss layouts in both node positions and bar cross-sectional areas.}
    \label{fig: Ablation 1}
\end{figure}

We present truss layouts only derived from {\FirstStage} and refined by {\SecondStage} separately in Fig. \ref{fig: Ablation 1}. In all cases, {\SecondStage} substantially reduces the total mass of the truss layout by $28\%$ on average, demonstrating the importance of {\SecondStage} in {\name} for further performance improvement.



\subsubsection{Usage of Diverse Truss Layouts}

To investigate the advantages of the diverse truss layouts derived in {\FirstStage}, we use the lightest truss layouts derived in {\FirstStage} without different topologies, named \textit{{\name} w.o. Diverse}. The results are presented in Tab. \ref{tab: Ablation_experiment_2}. {\name} outperforms \textit{{\name} w.o. Diverse} by an average of $3\%$ in all cases, which demonstrates the effectiveness of introducing diverse truss layouts.


\hide{We carry out a contrast experiment between the best truss layouts and diverse truss layouts as the input for {\SecondStage}. The former one uses top-$k$ best truss layouts as the initial states for {\SecondStage}, while the latter one uses $k$ different truss layouts, which is described in Sec. \ref{First Stage}. }

\hide{The ablation study is carried out in the 10-Bar test case, and we set $k = 5$ in practice. The results are shown in Tab. \ref{tab: Ablation_experiment_2}. Diverse truss layouts outperform single truss layouts by $3\%$ on average, revealing the fact that the best truss layout refined in {\SecondStage} may not come from the best truss layout in {\FirstStage}.}
    \begin{table}[!t]
		\centering
		\scriptsize
		\begin{tabular}{cccccccccccc}
		\toprule
		 Settings & {\name} w.o. Diverse & {\name} \\
		 \midrule
		 Load \uppercase\expandafter{\romannumeral1}, $p$ = 6 & 2149.60(1.90) & \textbf{2128.73(17.83)} \\
		 Load \uppercase\expandafter{\romannumeral2}, $p$ = 7 & 1419.67(18.45) & \textbf{1410.73(61.17)} \\
		 \bottomrule
		\end{tabular}
		\caption{Ablation studies on the usage of diverse truss layouts. \textit{{\name} w.o. Diverse} directly uses the lightest truss layouts derived in {\FirstStage} without different topologies. {\name} achieves better performance under all settings.} 
		\label{tab: Ablation_experiment_2}
    \end{table}

\subsubsection{Network Architecture}

Transformer and GNN architectures are commonly employed to handle graphical inputs. The comparison between \textit{GNN-based Policy} and {\name} is presented in Tab. \ref{tab: Ablation_experiment_3}. We utilized \verb|CGConv| \cite{Fey/Lenssen/2019} as GNN module, which has been demonstrated to exhibit good performance in material property prediction tasks \cite{xie2018crystal}.  The node positions, loads, and support information are embedded as nodes, and the cross-sectional area of each bar is recorded as an edge property. After GNN, we extract the embedding of the action node and then concatenate it with the action embedding for the final action. Empirically, we observe that a Transformer-based policy, as we used in {\name}, performs slightly better than a \textit{GNN-based Policy}. 

    \begin{table}[!t]
		\centering
		\scriptsize
		\begin{tabular}{cccccccccccc}
		\toprule
		 Settings & GNN-based Policy & {\name} \\
		 \midrule
		 Load \uppercase\expandafter{\romannumeral1}, $p$ = 6 & 2151.64(12.10) & \textbf{2128.73(17.83)} \\
		 Load \uppercase\expandafter{\romannumeral2}, $p$ = 7 & 1412.93(99.87) & \textbf{1410.73(61.17)} \\
		 \bottomrule
		\end{tabular}
		\caption{Ablation studies on the network architecture. {\name}, which adopts Transformer-based architecture, shows slightly better performance than \textit{GNN-based Policy}.}
		\label{tab: Ablation_experiment_3}
    \end{table}

\section{Conclusion}

We propose a two-stage method {\name} that can automatically design truss layouts under various constraints. We use UCT search to find diverse valid truss layouts in {\FirstStage} and then use deep RL policy to refine the truss layouts derived in {\FirstStage}. {\name} outperforms the baselines by $6.8\%$ on 2D testbed and $25.1\%$ on 3D testbed. {\name} may perform poorly when generating large-scale spatial structures, and combining basic structural elements in the search stage could accelerate the search speed. We leave this as our future work.


\section*{Acknowledgements}
The authors would like to acknowledge the financial support provided by the 2030 Innovation Megaprojects of China (Programme on New Generation Artificial Intelligence) Grant No. 2021AAA0150000 and Research on computer-generated intelligent design of buildings Grant No. SYXF0120020110.

\clearpage

\section*{Contribution Statement}
Authors Weihua Du and Jinglun Zhao contributed equally to this work and should be considered co-first authors.
\bibliographystyle{named}
\bibliography{ijcai23}

\appendix
\section{Appendix}
\subsection{Detailed Constraints}
\label{app: detailed constraints}
For any truss layout $G = (V, E)$, we have $8$ constraints in total to check whether the truss layout is valid. 

We define some parameters first: $\Omega$ is the design domain, $z_i$ is the cross-sectional area of $i$-th bar, $l_i$ is the length of $i$-th bar, $\sigma_i$ is the stress of $i$-th bar ($\sigma_i > 0$ means the bar is in tension and $< 0$ means in compression), $\delta_i$ is the displacement of $i$-th nodes from its original position after loaded, $\sigma_i^c = \max(0, -\sigma_i)$ is the compression part of stress, $b_i = \pi^2EI_i/z_il_i^2$ refers to the buckling limit of $i$-th bar ($E$ is Young's modulus, $I_i$ is the moment of inertia of $i$-th bar), $\lambda_i = l_i / \sqrt{I_i / z_i}$ is the slenderness ratio of $i$-th bar.

The $8$ constraints $g_0, ..., g_8$ are listed as following:

\begin{itemize}
\label{item: constraints}
\item Geometry stability$(g_0)$: the truss layout must pass three basic checks: (a) no extra degree of freedom, (b) stiffness matrix  $\succ 0$, (c) no intersection between bars;
\item Design domain$(g_1)$: each node and bar should be in the design domain $\Omega$;
\item Cross-sectional area$(g_2)$: each bar's cross-sectional area $z_i$ should be within $[z_{\min}, z_{\max}]$;
\item Stress constraint$(g_3)$: each bar's strength $\sigma_i$ should be within $[\sigma_{\min}, \sigma_{\max}]$;
\item Displacement$(g_4)$: each node's displacement $\delta_i$ should be within a small range in each direction. i.e, $\|\delta_i\|_\infty \leq \delta_{\max}$;
\item Stability$(g_5)$: each bar's compression part of stress $\sigma_i^c$ should be less than its buckle limit $b_i$.
\item Stiffness$(g_6)$: each bar's slenderness ratio $\lambda_i$ should be less than $\lambda_{\max}$;
\item Bar length$(g_7)$: each bar's length $l_i$ should be within $[l_{\min }, l_{\max }]$.
\end{itemize}

Notice that not all the constraints need to be satisfied in the test cases, and some test cases take the self-weight into consideration. We specify the constraint set in each test case in \ref{app: testbeds}.

\subsection{Detailed Settings of Testbeds}
\label{app: testbeds}
\subsubsection{10-Bar Cantilever Truss}
\label{app: 10-Bar Truss parameter}

The design domain of the 10-Bar Cantilever Truss is shown in Fig.~\ref{fig: 2Dtestbed} left. The 10-Bar Truss test case has $6$ fixed nodes. The left two nodes $(a, b)$ are support nodes, and the right four nodes $(c, d, e, f)$ may take some loads. The test case has two load cases. Load case \uppercase\expandafter{\romannumeral1} only has loads on nodes $(c, d)$, while Load case \uppercase\expandafter{\romannumeral2} has loads on all the four nodes $(c, d, e, f)$.  If a fixed node has no load and is not a support node, it will be removed from the initial truss layout. Constraints $\{g_0, g_1, g_2, g_3, g_4\}$ need to be satisfied in the 10-Bar test case. The node numbers $p$ of load case \uppercase\expandafter{\romannumeral1} and \uppercase\expandafter{\romannumeral2} are $6$ and $7$, respectively. This test case does not take the self-weight of bars into consideration.

Detailed information on fixed nodes is listed in Tab. \ref{tab: 10-Bar test case nodes}. Material properties and constraint parameter settings are listed in Tab. \ref{tab: 10-Bar test case constants}.
\begin{table}[!t]
\centering
\scriptsize
\begin{tabular}{cccccccccccc}
\toprule
Node & Location(mm) & Load Case 1 & Load Case 2\\
\midrule
a & (0, 0) & Support & Support \\
b & (0, 9144) & Support & Support \\
c & (9144, 0) & Loaded (0, -444,800 N) & Loaded (0, -667,200 N)\\
d & (18288, 0) & Loaded (0, -444,800 N) & Loaded (0, -667,200 N) \\
e & (9144, 9144) & N/A & Loaded (0, 444,800 N) \\
f & (18288, 9144) & N/A & Loaded (0, 444,800 N) \\
\bottomrule
\end{tabular}
\caption{Each fixed node information of the 10-Bar test case.}
\label{tab: 10-Bar test case nodes}
\end{table}
\begin{table}[!t]
\centering
\scriptsize
\begin{tabular}{cccccccccccc}
\toprule
Parameters & Values \\
\midrule
Design Domain$(\Omega)$ & $[0, 18,288]\times [0, 9,144]$ mm \\
Young’s modulus$(E)$ & $68,950 \text{ MPa} (10,000 \text{ ksi})$ \\
Density$(\rho)$ & $2767.99 \text{ kg/m}^3 (0.1 \text{ lb/in}^3)$ \\
Stress range$(\sigma_{\min}, \sigma_{\max})$ & $[-172.369, 172.369] \text{ MPa} ([-25, 25] \text{ksi})$ \\
Max node displacement$(\delta_{\max})$ & $50.8 \text{ mm} (2 \text{ in})$ \\
Bar area range$(z_{\min}, z_{\max})$ & $[0.6452, 225.806]\text{ cm}^2$ \\
Consider self-weight$(f_{self})$ & No\\
\bottomrule
\end{tabular}
\caption{Detailed constant information of the 10-Bar test case.}
\label{tab: 10-Bar test case constants}
\end{table}

\subsubsection{17-Bar Cantilever Truss}
\label{app: 17-Bar Truss parameter}

The design domain of the 17-Bar Cantilever Truss is shown in Fig.~\ref{fig: 2Dtestbed} right. The 17-Bar Truss test case has $3$ fixed nodes. The left two nodes $(a, b)$ are support nodes, and node $(i)$ takes some loads. Constraints $\{g_0, g_1, g_2, g_3, g_4, g_5\}$ need to be satisfied in the 17-Bar test case. The node number $p$ is $6$. This test case does not take the self-weight of bars into consideration.

Detailed information on fixed nodes is listed in Tab. \ref{tab: 17-Bar Truss nodes}. Material properties and constraint parameter settings are listed in Tab. \ref{tab: 17-Bar Truss constant}.
\begin{table}[!t]
\centering
\scriptsize
\begin{tabular}{cccccccccccc}
\toprule
Node & Location(mm) & Load \\
\midrule
a & (0.0, 0.0) & Support \\
b & (0.0, 2540.0) & Support \\
i & (10160.0, 0.0) & Loaded (0, -444,800 N)  \\
\bottomrule
\end{tabular}
\caption{Each fixed node information of the 17-Bar test case.}
\label{tab: 17-Bar Truss nodes}
    \end{table}
    \begin{table}[!t]
\centering
\scriptsize
\begin{tabular}{cccccccccccc}
\toprule
Parameters & Values \\
\midrule
Design Domain$(\Omega)$ & $[0, 10,160]\times [0, 2,540]$ mm \\
Young’s modulus$(E)$ & $206,850 \text{ MPa} (30,000 \text{ ksi})$\\
Density$(\rho)$ & $7418.21 \text{ kg/m}^3 (0.268 \text{ lb/in}^3)$ \\
Stress range$(\sigma_{\min}, \sigma_{\max})$ & $[-334.6, 334.6] \text{ MPa} ([-50, 50] \text{ksi})$ \\
Max node displacement$(\delta_{\max})$ & $50.8 \text{ mm} (2 \text{ in.})$ \\
Bar area range$(z_{\min}, z_{\max})$ & $[0.6452,225.806]\text{ cm}^2$ \\
Consider self-weight$(f_{self})$ & No\\
\bottomrule
\end{tabular}
\caption{Detailed constant information of the 17-Bar test case.}
\label{tab: 17-Bar Truss constant}
\end{table}
\subsubsection{3D Cantilever Sundial}
\label{app: 3D setting parameter}

The design domain of the 3D Cantilever Sundial test case is shown in Fig.~\ref{figure: 3Dtestbed}. The design domain only represents node locations and there is no mandatory geometric boundary for newly added nodes and bars. There are four fixed nodes in the design domain, among which nodes $(1, 2, 3)$ are support nodes, and node $(4)$ is the sundial tip with load $50$ N. Constraints $\{g_0, g_1, g_2, g_3, g_4, g_6, g_7\}$ need to be satisfied in the Sundial test case. The node number $p$ is $7$ or $8$ or $9$. This test case takes the self-weight of bars into consideration.

Detailed information on fixed nodes is listed in Tab. \ref{tab: 3D setting nodes}. Material properties and constraint parameter settings are listed in Tab. \ref{tab: 3D setting constant}. Note that the slenderness ratio limit is different for tension bars and compression bars ($220$ for tension bars and $180$ for compression bars).

The cross-section of the bars used in this section is the section of cold-formed thin-wall welded round steel tube (GB50018-2002)\cite{gb500182002technical}. There are $61$ kinds of cross-sections in total and each area $z$ of the cross-sections is defined by diameter $d$ and thickness $t$: $z = \pi(d^2 / 4 + (d - 2t)^2 / 4)$, with parameters ranging from $d~25~t~1.5$ to $d~245~t~4.0$.
\begin{table}[!t]
\centering
\scriptsize
\begin{tabular}{cccccccccccc}
\toprule
Node & Location(mm) & Load \\
\midrule
a & (0.0, 0.0, 0.0) & Support \\
b & (0.0, -483, 595) & Support \\
c & (0.0, 483, 595) & Support\\
d & (4634, 772, -78) & Loaded (0, 0, -50 N) \\
\bottomrule
\end{tabular}
\caption{Each fixed node information of the 3D Sundial test case.}
\label{tab: 3D setting nodes}
    \end{table}
    \begin{table}[!t]
\centering
\scriptsize
\begin{tabular}{cccccccccccc}
\toprule
Parameters & Values \\
\midrule
Design Domain$(\Omega)$ & No mandatory geometric boundary \\
Young’s modulus$(E)$ & $193 \text{ GPa}$ \\
Density$(\rho)$ & $8000 \text{ kg/m}^3 (0.268 \text{ lb/in}^3)$ \\
Strength range & $[-123, 123] \text{ MPa}$ \\
Max node displacement$(\delta_{\max})$ & $2 \text{ mm}$ \\
Slenderness ratio$(\lambda_{\max})$ & $220$(tension bar) and $180$(compression bar)\\
Bar length range$(l_{\min}, l_{\max})$ & $[0.03, 5]$ m\\
Bar area range $(z)$ & Cross-sections in GB50018-2002\\
Consider self-weight$(f_{self})$ & Yes\\
\bottomrule
\end{tabular}
\caption{Detailed constant information of the 3D Sundial test case.}
\label{tab: 3D setting constant}
\end{table}
\subsection{Detailed Baseline Description}
\label{app: baseline}
\subsubsection{AlphaTruss}
\textit{AlphaTruss}\cite{luo2022alphatruss} is a two-stage search method. The first stage is searching in discrete space by UCT search. Similar to \textit{AutoTruss}, it searches for node position, node connection, and cross-sectional area of bars sequentially. The second stage is used to refine the best truss layouts generated in the first stage by discretized UCT search, too. In detail, suppose $w$ is the step size of the discretization in the first stage and $p$ is the search result of a node's position or a bar's cross-sectional area. The search domain will be restricted to $[p - w / 2, p + w / 2]$.

\subsubsection{KR-UCT}
\textit{KR-UCT}\cite{luo2022reinforcement} is a one-stage search method, which uses UCT search directly. To handle the continuous search space, it applies the kernel method. Similarly to \textit{AlphaTruss}, it searches node position, node connection, and cross-sectional area of bars sequentially. To the best of our knowledge, it is the first search method that can apply to 3D settings without any predefined structure.

\subsubsection{SEOIGE}
\textit{SEOIGE}\cite{fenton2015discrete} uses grammatical evolution to represent a variable number of nodes and positions on a continuum, and then uses the Delaunay triangulation algorithm to build bars between nodes. \textit{SEOIGE} works well in test cases where the structure of the solution is not known a priori.

\subsection{Network Architecture}
\label{app: network details}
All in all, the network architecture of the RL policy for \textit{AutoTruss} has four parts: node/bar/action id/action embedding, self-attention encoder, action decoder, and action/value head.
\subsubsection{Node/Bar/Action id/Action Embedding}
For a truss layout $G = (V, E)$, \textit{AutoTruss} first embeds each node $v_i$ and bar $e_i$ by MLPs. In detail, a node $v_i$ is represented by [\text{position}, \text{ support condition}, \text{ load condition}] and put into a two-layer MLP with hidden dim $128$ and output dim $256$. Similarly, a bar $e_i$ is represented by [\text{node position } 1, \text{node position }2, \text{cross-sectional area}] and put into a two-layer MLP with the same hidden dim and output dim as node embedding. For the action id and action, we also use two two-layer MLPs with the same hidden dim and output dim. Let the embedded nodes, bars, action id, and action be $\hat{v_1}, ..., \hat{v_{|V|}}$, $\hat{e_1}, ..., \hat{e_{|E|}}$, $\hat{id}$, and $\hat{a}$, respectively. 

\subsubsection{Self-Attention Encoder}
To get the connection between nodes, bars, and action id of a truss layout. We use a self-attention encoder to extract information. First, we concatenate the embedding of nodes and bars as a sequence $[\hat{v_1}, ..., \hat{v_{|V|}}, \hat{e_1}, ..., \hat{e_{|E|}}]$, and then put the sequence into the self-attention encoder. The self-attention encoder has hidden dim $256$ and $6$ layers.

\subsubsection{Action Decoder}
To get the predicted action $a$ and the Q value $Q(s, a)$. Action id $\hat{id}$ and action $\hat{a}$ are put into the action decoder, which is a $6$ layer decoder with hidden dim $256$. Let the hidden state of action id and action generated by action decoder be $h_{id}$ and $h_{a}$, respectively.

\subsubsection{Action/Value Head}
We use the hidden state of action id $h_{id}$ and action $h_{a}$ to generate action $a$ and predict Q-value $Q(s, a)$, respectively. To generate action $a$, we put $h_{id}$ into a three-layer MLP with hidden dims $256, 512$. Similarly, to predict Q-value $Q(s, a)$, we put $h_{a}$ into another MLP with the same hidden dims.

\subsection{Hyperparameters}
\label{app: hyperparameters}
There are $6$\hide{\jz{5 hyperparameters?}} hyperparameters in our algorithm, namely the first exploration parameter $\beta$ in Equ.~(2), the second exploration parameter $c$ in Equ.~(3), discount factor $\gamma$ in Equ.~(5), temperature parameter $\alpha$ in Equ.~(6), and the learning rate of policy and Q-value function $lr_{policy}$ and $lr_{Qf}$. \hide{and critic parameter $\psi$ in Equ.~(7)}. The values of the hyperparameters are chosen through trial and error, with their values listed in Tab.~\ref{tab: hyperparameters}.
\begin{table}[!t]
\centering
\scriptsize
\begin{tabular}{cccccccccccc}
\toprule
Hyperparameters & Values \\
\midrule
Exploration parameter I$(\beta)$ & $0.3$ \\
Exploration parameter II$(c)$ & $30$\\
\midrule
Discount factor$(\gamma)$ & $0.99$ \\
Initial temperature$(\alpha)$ & $1.0$ \\
Policy learning rate$(lr_{policy})$ & $0.0003$ \\
Q-value learning rate$(lr_{Qf})$ & $0.0003$ \\
\bottomrule
\end{tabular}
\caption{Hyperparameters used in UCT, RL, and SAC.}
\label{tab: hyperparameters}
\end{table}

\subsection{Data of Generated Truss Layouts}
\begin{figure}[!t]
\centering
    \includegraphics[width=1\linewidth]{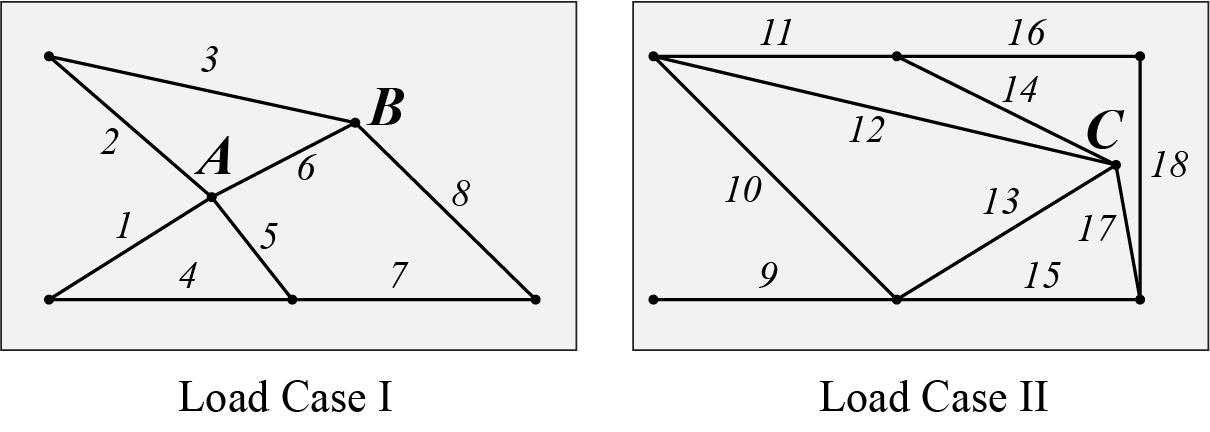}
    \caption{Illustration for the generated truss in 2D 10-bar.}
    \label{figure: detailed 10 bar}.
\end{figure}

\begin{figure}[!t]
\centering
    \includegraphics[width=0.8\linewidth]{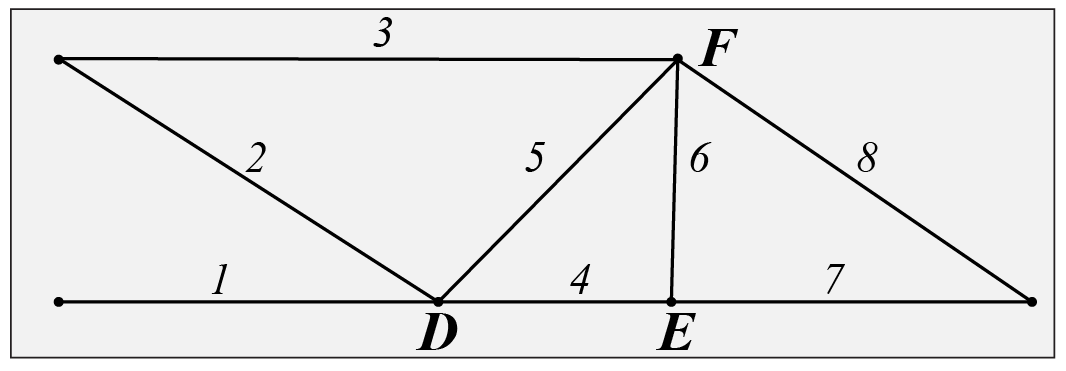}
    \caption{Illustration for the generated truss in 2D 17-bar.}
    \label{figure: detailed 17 bar}.
\end{figure}

Detailed data of the generated truss layouts, including node coordinates and cross-sectional areas of the bars are provided in tables. Tab.~\ref{tab: detail 10 bar} shows the details for both load case \uppercase\expandafter{\romannumeral1} and \uppercase\expandafter{\romannumeral2} for the 10-bar test case with illustration in Fig.~\ref{figure: detailed 10 bar}. Tab.~\ref{tab: detail 17 bar} shows the details for the 17-bar test case with illustration in Fig.~\ref{figure: detailed 17 bar}.  Tab.~\ref{tab: detail 3D node} and Tab.~\ref{tab: detail 3D bar} show the details for the 3D test case.

\begin{table}[!t]
\centering
\scriptsize
\begin{tabular}{ccccccccccc}
\toprule
Node Label & \multicolumn{2}{c}{A} & \multicolumn{2}{c}{B} & \multicolumn{2}{c}{C}\\
\midrule
Coord. (mm) & \multicolumn{2}{c}{(6115,3851)} & \multicolumn{2}{c}{(11508,6647)} & \multicolumn{2}{c}{(17380,5062)}\\
\midrule
& 1 & 2 & 3 & 4 & 5 & 6\\
& 125 & 31.8 & 126 & 189 & 33.2 & 117\\
Bar Label& 7 & 8 & 9 & 10 & 11 & 12\\
Area (cm$^2$) & 97.9 & 140 & 165 & 88.7 & 53.6 & 1.06\\
& 13 & 14 & 15 & 16 & 17 & 18\\
& 58.4 & 57.0 & 14.0 & 0.65 & 63.1 & 12.9\\
\bottomrule
\end{tabular}
\caption{ Detailed data of the generated truss in 2D 10-bar.}
\label{tab: detail 10 bar}
\end{table}

\begin{table}[!t]
\centering
\scriptsize
\begin{tabular}{ccccccccc}
\toprule
Node Label & \multicolumn{3}{c}{D} & \multicolumn{2}{c}{E} & \multicolumn{3}{c}{F}\\
\midrule
Coord. (mm) & \multicolumn{3}{c}{(3963,0)} & \multicolumn{2}{c}{(6395,0)} & \multicolumn{3}{c}{(6462,2528)}\\
\midrule
Bar Label & 1 & 2 & 3 & 4 & 5 & 6 & 7 & 8\\
Area (cm$^2$) & 132 & 38.6 & 47.6 & 48.7 & 71.4 & 0.65 & 76.1 & 41.4\\
\bottomrule
\end{tabular}
\caption{ Detailed data of the generated truss in 2D 17 bar.}
\label{tab: detail 17 bar}
\end{table}

\begin{table}[!t]
\centering
\scriptsize
\begin{tabular}{cccc}
\toprule
&\multicolumn{3}{c}{3D Case Node Coordinates (m)}\\
\# & 7-Point & 8-Point & 9-Point\\
\midrule
1 & (0, 0, 0) & (0, 0, 0) & (0, 0, 0)\\
2 & (0, -0.48, 0.6) & (0, -0.48, 0.6) & (0, -0.48, 0.6)\\
3 & (0, 0.48, 0.6) & (0, 0.48, 0.6) & (0, 0.48, 0.6)\\
4 & (1.66, 0.24, -0.11) & (4.63, 0.77, -0.08) & (4.63, 0.77, -0.08)\\
5 & (1.69, 0.3, 0.18)& (1.67, 0.31, -0.03) & (2.29, 0.38, -0.12)\\
6 & (1.64, 0.22, 0.27) & (1.77, 0.37, 0.34) & (1.57, 0.27, 0.22)\\
7 & (4.63, 0.77, -0.08) & (0.43, -0.08, 0.43) & (0.6, 0.14, 0.57)\\
8 & & (1.72, 0.29, 0.38) & (1.74, 0.36, 0.24)\\
9 & &&(1.71, 0.23, 0.22)\\
\bottomrule
\end{tabular}
\caption{ Node coordinates detail of the generated truss in 3D.}
\label{tab: detail 3D node}
\end{table}

\begin{table}[!t]
\centering
\scriptsize
\begin{tabular}{cccccccccccc}
\toprule
\multicolumn{6}{c}{Bar Connection, Outer Diameter(mm)}\\
\multicolumn{6}{c}{(Thickness=1.5 mm, if not stated otherwise)}\\
\multicolumn{2}{c}{7-Point} & \multicolumn{2}{c}{8-Point} & \multicolumn{2}{c}{9-Point}\\
\midrule
1-5: 30.0 & 3-5: 25.0 & 1-5: 30.0 & 3-5: 25.0 & 4-9: 40.0 & 1-4: 40.0\\
1-6: 30.0 & 3-6: 25.0 & 3-6: 25.0 & 4-6: 40.0 & 3-5: 25.0 & 4-5: 25.0\\
4-6: 40.0 & 5-6: 25.0 & 5-6: 25.0 & 1-7: 25.0 & 1-6: 25.0 & 2-6: 25.0\\
1-7: 30.0 & 2-7: 25.0 & 2-7: 25.0 & 3-7: 25.0 & 3-6: 25.0 & 7-9: 40.0\\
3-7: 25.0 & 5-7: 25.0 & 5-7: 25.0 & 1-8: 30.0 & 1-7: 25.0 & 4-7: 25.0\\
6-7: 25.0 &  & 3-8: 25.0 & 4-8: 40.0 & 5-7: 25.0 & 6-7: 25.0\\
\multicolumn{2}{c}{4-5: 51.0 $\leftarrow$ 2mm thick} & 6-8: 25.0 & 7-8: 25.0 & 8-9: 40.0 & 1-8: 25.0\\
\multicolumn{2}{c}{4-7: 51.0 $\leftarrow$ 2mm thick} & \multicolumn{2}{c}{4-7: 51.0 $\leftarrow$ 2mm thick} & 3-8: 30.0 & 4-8: 25.0\\
 &  &  &  & 5-8: 25.0 & 6-8: 25.0\\
 &  &  &  & 7-8: 25.0 & \\
\bottomrule
\end{tabular}
\caption{ Bar connections, outer diameter and thickness of the generated truss in 3D.}
\label{tab: detail 3D bar}
\end{table}


\subsection{Ablation Study on Environment}

To check the design of the RL environment, we conduct an ablation study on whether the environment allows invalid truss layouts during rollout. Under our current design, the environment allows the policy generates invalid truss layouts in less than $5$ steps within a single episode. Another environment design does not allow the policy generates any invalid layouts, and it will stop the episode immediately if any invalid layout occurs, named \textit{AutoTruss w.o Invalid}. We run the ablation study on the 10-Bar test case for $3$ seeds, and the comparison is shown in Tab. \ref{tab: Ablation Environment}. Data is reported as "mean(standard deviation)". Allowing the occurrence of invalid truss layouts achieves better results since it encourages the RL policy to explore more.
\begin{table}[!t]
    \centering
    \scriptsize
    \begin{tabular}{cccccccccccc}
    \toprule
     Settings & {\name} w.o. Invalid & {\name} \\
     \midrule
     Load \uppercase\expandafter{\romannumeral1}, $p$ = 6 & 2136.01\tiny(14.91) & \textbf{2128.73\tiny(17.83)} \\
     Load \uppercase\expandafter{\romannumeral2}, $p$ = 7 & 1631.41\tiny(128.69) & \textbf{1410.73\tiny(61.17)} \\
     \bottomrule
    \end{tabular}
    \caption{Ablation studies on whether the environment allows invalid truss layouts. \textit{{\name} w.o. Invalid} does not allow any invalid truss layouts to occur in the rollout. {\name} achieves better performance under all settings.} 
    \label{tab: Ablation Environment}
\end{table}

\subsection{Comparison of Running Time}
\label{app: running time}
The parameters for the number of episodes used in the UCT search and the number of environment steps used in the reinforcement learning were chosen to ensure a fair comparison to other baselines, by keeping a similar time cost.

Our first baseline, \textit{KR-UCT}, uses a single stage of searching with an upper limit of $4$ million iterations when running the 3d kr-sundial test case. In contrast, since our algorithm has two stages, we set the upper limit of the search stage to $2$ million iterations to keep the search time half that of \textit{KR-UCT}. $150,000$ environment steps ensure that the refinement stage takes the same time as the search stage.

The second baseline, \textit{AlphaTruss}, also has two stages, and its search stage takes approximately half the number of iterations as \textit{KR-UCT}. Therefore, the computation for our search stage is consistent with \textit{AlphaTruss}. 

We report the number from their paper for the third baseline \textit{SEOIGE} since its codebase is not public. 

\newpage

\end{document}